\documentclass[10pt,twocolumn,letterpaper]{article}

\usepackage{graphicx}
\usepackage{amsmath}
\usepackage{amssymb}
\usepackage{booktabs}
\usepackage{array}

\usepackage[pagebackref,breaklinks,colorlinks]{hyperref}
\usepackage{multirow}

\usepackage[capitalize]{cleveref}
\crefname{section}{Sec.}{Secs.}
\Crefname{section}{Section}{Sections}
\Crefname{table}{Table}{Tables}
\crefname{table}{Tab.}{Tabs.}

\begin{document}

\title{ HBReID: Harder Batch for Re-identification}

\author{Wen Li,
Furong Xu, 
Jianan Zhao, 
Ruobing Zheng\\
Cheng Zou,
Meng Wang,
Yuan Cheng\\
{\tt\small }
ANT GROUP\\
{\tt\small \{yinian.lw,booyoungxu.xfr,zhaojianan.zjn,zhengruobing.zrb}\\
{\tt\small wuyou.zc,darren.wm,chengyuan.c\}@antgroup.com }
}

\date{}

\maketitle

\begin{abstract}
Triplet loss is a widely adopted loss function in ReID task which pulls the hardest positive pairs close and pushes the hardest negative pairs far away. However, the selected samples are not the hardest globally, but the hardest only in a mini-batch, which will affect the performance. In this report, a hard batch mining method is proposed to mine the hardest samples globally to make triplet harder. More specifically, the most similar classes are selected into a same mini-batch so that the similar classes could be pushed further away. Besides, an adversarial scene removal module composed of a scene classifier and an
adversarial loss is used to learn scene invariant feature representations. Experiments are conducted on dataset MSMT17 to prove the effectiveness, and our method surpasses all of the previous methods and sets state-of-the-art result. 

\end{abstract}

\section{Introduction}
\label{sec:intro}

Object re-identification (ReID) aims to associate a particular object across different scenes and camera views, such as in the applications of person ReID and vehicle ReID. Academic research is progressing in four major directions: image preprocessing, feature representation learning, metric learning, and ranking optimization~\cite{Zhong_2017_CVPR,zhang2021graph}.

As for image preprocessing, data augmentation is crucial to ReID. Random cropping, padding, horizontal flipping and random erasing~\cite{RandomErasing} are frequently-used methods which will increase the number and diversity of training samples. Grayscale patch replacement~\cite{GS} is proposed to randomly replace a rectangular region of RGB pixels in the image with its corresponding grayscale counterpart, which has been proved effective. 

As for representation learning, CNN-based~\cite{Yang_2021_ICCV,Khorramshahi2020TheDI,Sun_2020_CVPR,Wang_2020_CVPR,Isobe_2021_ICCV} architecture has dominated for a long time until transformer-based~\cite{liang2021cmtr,chen2021ohformer} methods appear. TransReID~\cite{He_2021_ICCV}, a pure transformer method, has achieved state-of-the-art on many ReID benchmarks. 

Metric learning often concentrates on designing powerful loss functions to extract robust feature embeddings. In addition to Softmax-cross-entropy, proxy-based loss and pair-based loss~\cite{HermansBeyer2017Arxiv,Kim_2020_CVPR,Wang_2019_CVPR_Multi-Similarity,Wang_2019_CVPR_Ranked,Wei_2018_CVPR} are used to pull the same classes close and push the different classes far away. Some methods also use domain adaption to reduce domain gap for retrieving under different cameras.

\begin{figure*}[t]
\begin{center}
\includegraphics[scale=0.5]{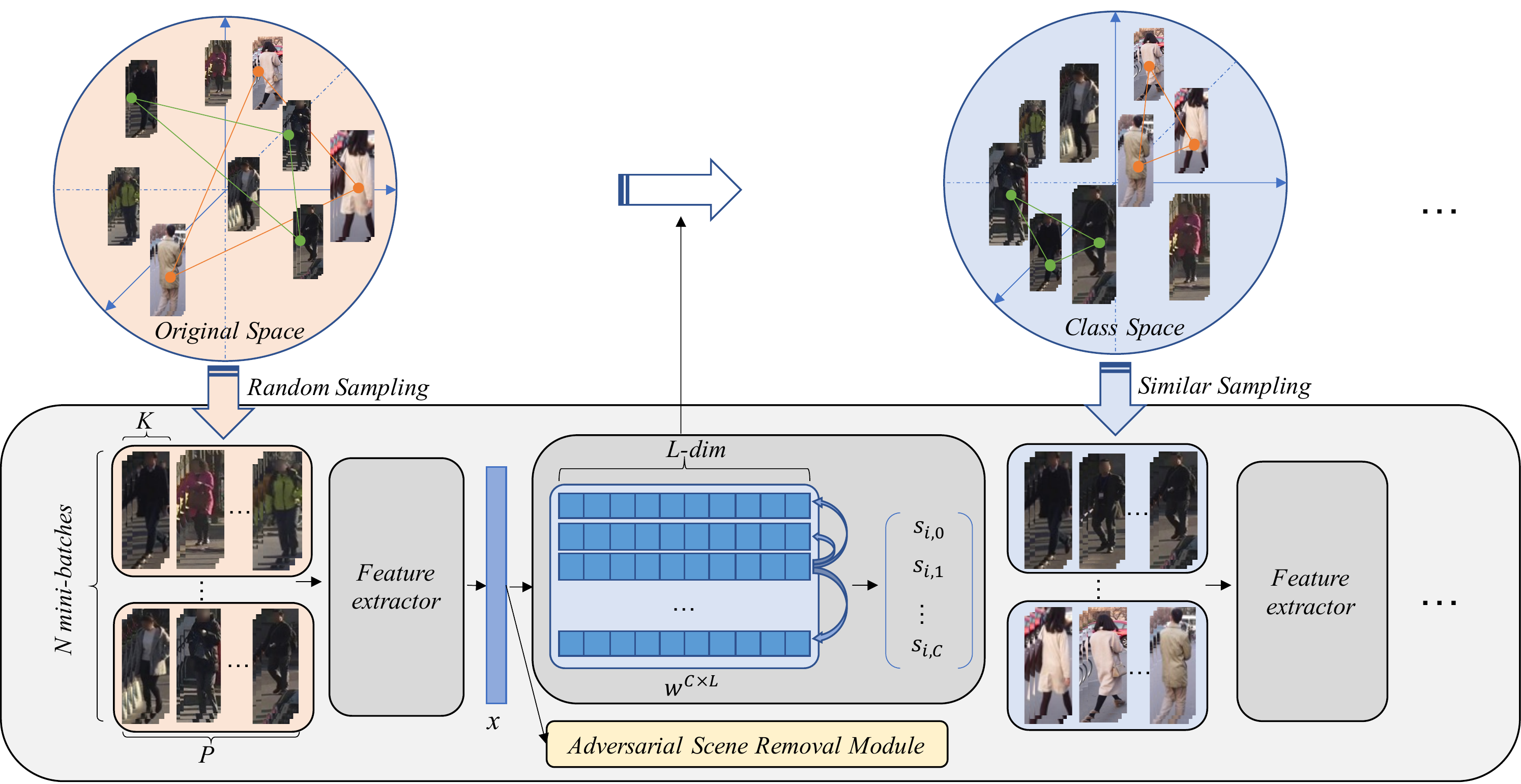}
\end{center}
\caption{The main framework of HBReID. In the first few steps, it randomly sample $P$ person to construct mini-batches. And then in the following steps, it samples the most similar classes given a random class ID according to the class similarity measured by the learnt class weights, then updates the model. This process is called hard batch mining, and it will iteratively run until the training is finish. Note that there is an adversarial scene removal module on top of the global feature $x$. $ w^{C \times L}$ is the class weights of the last fully connected layer.}
\label{fig:framework}
\end{figure*}

In this report, we propose hard batch mining for ReID which mines the hardest samples globally to make triplet harder. Besides, an adversarial scene removal module composed of a scene classifier and an adversarial loss is used to learn scene invariant feature representations. The contributions are summarized as follows:
\begin{itemize}
\item We propose a hard batch mining method to mine the hardest samples globally to make triplet harder during training.
\item We propose an adversarial scene removal module to learn scene invariant feature representations.
\item Our method surpasses all of the previous methods and sets state-of-the-art result on MSMT17.
\end{itemize}

\section{Related Work}
\label{sec:formatting}
\subsection{Triplet Loss}
In classification task and metric learning, Softmax-cross-entropy is a mainstream loss function to split the different class into different feature space. But it can't handle far intra-class distance and close inter-class distance very well. Therefore, proxy-based loss~\cite{teh2020proxynca++,Kim_2020_CVPR} and pair-based loss~\cite{Wu_2017_ICCV,Ustinova_2016_NIPs} are used to pull the samples of the same class closer and to push the samples of different classes far away. Triplet loss 
is the most prominent approach that uses pair-based distanced calculation for deep metric learning~\cite{Wu_2017_ICCV}. During training, a sample in a mini-batch is considered as an anchor, the distances of negative samples and positive samples with the anchor are calculated to select the closest negative samples and the furthest positive samples. Triplet loss minimizes the distance of the anchor-positive pairs and tries to maximize that of the anchor-negative pairs. But in most cases, the selected samples for triplet loss are not the hardest globally, but the hardest only in a mini-batch, which is suboptimal.

\subsection{Domain Adaption}

Domain adaption aims at learning from a source data distribution and well performing on a different (but related) target data distribution~\cite{kouw2019introduction}. ReID can be regarded as a domain-variant problem because it retrieves the same person from diverse cameras with various illuminations and viewpoints which are different at training and test phase. Learning domain-invariant feature representations to reduce the distance of different domains is a common method~\cite{NIPS2006_Gretton,sun2016correlation,1997On_Information_Sufficiency,Kang_2019_CVPR}. Re-weighting the instances selected from a subset of two domains where they have smaller distance is also useful. Adversarial networks~\cite{ganin2016domainadversarial,Bousmalis_2017_CVPR,cao2018partial,Tzeng_2017_CVPR} or simply an adversarial loss~\cite{Gebru_2017_ICCV,Tzeng_2015_ICCV} are also used to confuse networks to focus more on domain-invariant features.

\section{Proposed Method}
\label{sec:formatting}

The proposed method is based on TransReID~\cite{He_2021_ICCV}. To further improve the performance, a hard batch mining method is proposed to sample closer classes to construct harder mini-batches during training. Meanwhile, an adversarial scene removal module is used to make the feature representations scene invariant. Figure.~\ref{fig:framework} shows the main framework of the proposed method.

\subsection{Hard Batch Mining}

PK sampling is an effective sampling strategy in ReID. P classes are randomly selected at first, and then K images are randomly selected from each class, so that there are PK images in a mini-batch. For each sample in the mini-batch, the hardest positive and the hardest negative samples can be selected for triplet loss. However, the similarities between classes can be quite different. Some classes are extremely similar while some are obviously dissimilar. Random sampling P classes per mini-batch directly is a limitation for further hard mining. In this report, we sample P classes with the highest similarity so that the hardest anchor-negative pairs can be mined for triplet loss to improve the discrimination of similar classes.

Inspired by DAM~\cite{Xu_2021_CVPR}, the weights $w$ ($w \in R ^{C \times L}$, where $C$ is class number, and $L$ is the length of classified embedding) of the last fully connected layer are used to get the pairwise similarity between classes. Mathematically, a feature $x_i$ is projected onto all weight vectors $[w_1 , ..., w_C ]$ to determine its class, where $w_i$ is a $L$-dim vector. Therefore, class weights can be used to represent the average feature of intra-class samples (class center), and the similarity between class weights can represent the distance between inter-classes. The class similarity $s_i,_j$ between class $c_i$ and $c_j$ can be defined as 

\begin{equation}
  s_i,_j = cos(w_i, w_j)
  \label{eq:important}
\end{equation}

The higher the value is, the more similar two classes are. The common approach is to select the most difficult samples from the current remaining data in each iteration to form a batch, but multiple similarity calculations affect the training speed. Inspired by the equidistant constraint in EET~\cite{Xu_2020_WWW}, when some samples have the closed same distance from a certain sample, these samples are always distributed closer. Therefore, for each epoch of training, we randomly select a class to calculate the similarity with other classes. Then all classes are sorted according to the similarity, and the sorted classes are constructed batch in turn for training. This batch sampling operation allows classes that are closer to each other to be optimized in a mini-batch, thereby improving the distance distribution between classes.

\begin{figure}[t]
\begin{center}
\includegraphics[scale=0.35]{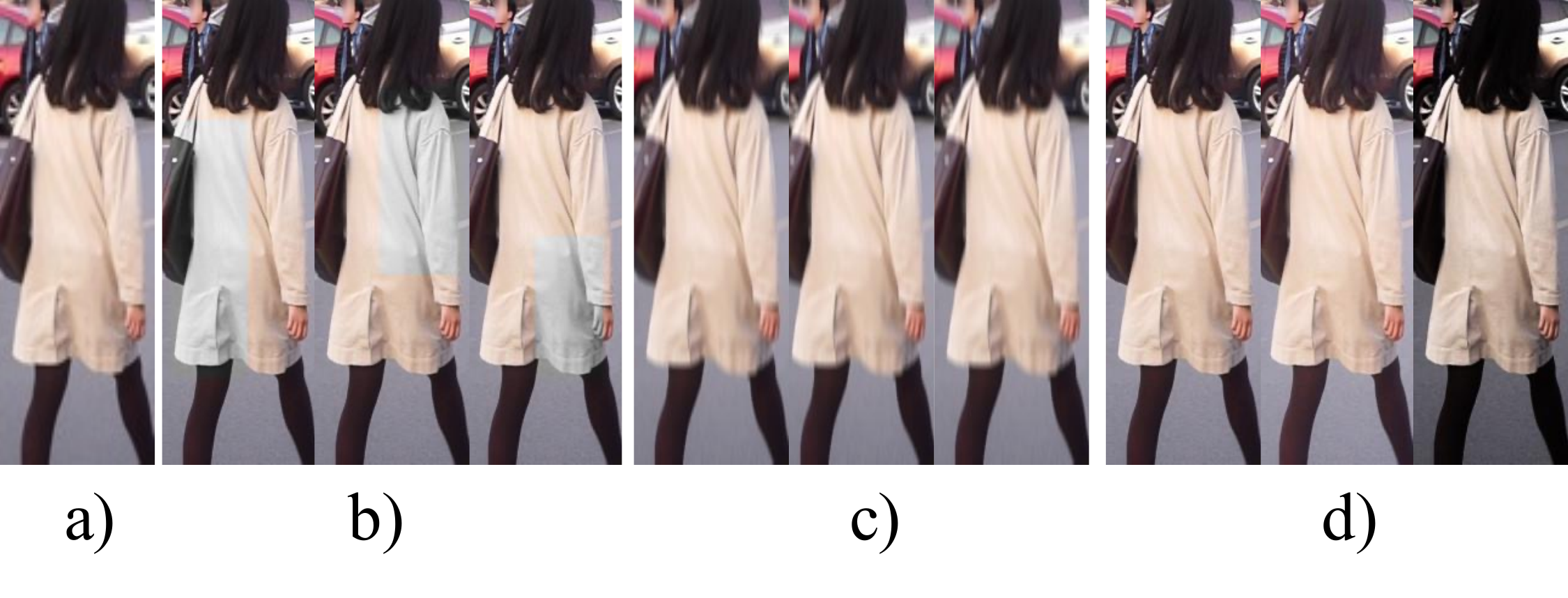}
\caption{Different augmentation examples. a) Original. b) Grayscale patch replacement. c) Gaussian blur. d) Gamma transformation.}
\label{fig:augmentation}
\end{center}
\end{figure}

\subsection{Adversarial Scene Removal}

ReID task is to retrieve the same person in diverse cameras which are placed at different scenes with various illuminations and viewpoints. To remove the influence of scene types, a classifier with adversarial loss is designed following~\cite{Choi2019WhyCI}. The motivation behind this scene removal is that we want to learn feature representations for re-identification but invariant to different scene types. We learn a scene classifier $f_s$ on top of the global feature $x$ in an adversarial way. The adversarial loss $L_{adv}$ is defined as
\begin{equation}
  L_{adv} = -\sum_{t=1}^{T} y_{t} log(f_s(x))
  \label{eq:important}
\end{equation} 

where $T$ is the number of scene types and $y_t$ is scene label. The loss is adversarial because the scene classifier aims to minimize it, while the feature extractor tries to maximize it to confuse the classifier. In the end, one can't tell the scene types from the feature, that is, the feature is scene invariant.

The total loss is then defined as
\begin{equation}
  L_{total} = L_{ReID} - \lambda L_{adv}
  \label{eq:important}
\end{equation} 

where $\lambda$ is a positive coefficient to balance the scene adversarial loss.



\begin{table}[t]
  \caption{Performance on MSMT17 dataset with different data augmentations. CJ: color jitter, GB: gaussian blur, GT: gamma transformation, GS: grayscale patch replacement. \\}
  \label{tab:augmentation}
  \centering
  \begin{tabular}{|c c c c c|}
    \toprule
    Default & +CJ & +GB & +GT & +GS \\
    \midrule
    68.8 & -3.5 & -0.6 & -0.3 & +0.8 \\
    \bottomrule
  \end{tabular}
  \centering
\end{table}\

\section{Experiment}
\label{sec:formatting}

\subsection{Implementation Details}

The experiments are conducted on person ReID dataset MSMT17~\cite{Wei_2018_CVPR}. All images are resized to 384$\times$128. The batch size is set to 64 with 4 images per ID. SGD optimizer is employed with a momentum of 0.9 and the weight decay of 1e-4. The learning rate is initialized as 0.008 with cosine learning rate decay. All the experiments are performed with 4 Nvidia Tesla V100 GPUs. The initial weights of ViT are pre-trained on ImageNet-21K and then finetuned on ImageNet-1K. We evaluate all methods with Cumulative Matching Characteristic (CMC) curves and the mean Average Precision (mAP).

\subsection{Data Augmentation}
Data augmentation is crucial to ReID tasks. Proper augmentations can improve performance while improper ones could make it worse. We use the augmentations including horizontal flipping, padding, random crop and random erasing which are commonly used in ReID. Besides, another augmentation, grayscale patch replacement~\cite{GS}, is used to randomly select a rectangle region in an image and $grayscale$ it, see Figure.~\ref{fig:augmentation}. Experiments of other color distortion augmentations like color jitter, gaussian blur and gamma transform are reported in Table.~\ref{tab:augmentation}, and the result indicates that grayscale patch replacement is effective.

\begin{table}[t]
  \caption{Ablation studies and comparisons with SOTA methods. $\S$ indicates our re-implementation, $\dagger$ indicates test with re-ranking~\cite{Zhong_2017_CVPR}. GS: grayscale patch replacement, HM: hard batch mining, SR: adversarial scene removal module. \\}
  \label{tab:sota}
  \centering
  \begin{tabular}{|c m{0.7cm} m{0.7cm} m{0.7cm} m{0.7cm} m{0.7cm}|}
    \toprule
    Method & \  & \  & \  & $mAP$ & $cmc1$ \\
    \midrule
    FlipReID~\cite{ni2021flipreid} & \  & \  & \  & 68.0 & 85.6 \\
    TransReID~\cite{He_2021_ICCV} & \  & \  & \  & 69.4 & 86.2 \\
    FlipReID$\dagger$ & \  & \  & \  & 81.3 & 87.5 \\
    TransReID$\S\dagger$ & \  & \  & \  & 83.0 & 88.8 \\
    \hline
    \ & +GS & +HM & +SR  & \  & \ \\
    \hline
    TransReID\S & \  & \  & \  & 68.8 & 85.8 \\
    \hline
    \multirow{4}{*}{ours} & $\surd$ & \  & \  & 69.6 & 86.3 \\
    & $\surd$ & $\surd$ & \  & 70.1 & 86.9 \\
    & $\surd$  & \  & $\surd$  & 69.8 & 86.9 \\
    & $\surd$  & $\surd$  & $\surd$  & 70.1 & 87.0 \\
    \hline
    ours$\dagger$ & $\surd$  & $\surd$  & $\surd$  & 84.4 & 89.9 \\
    \bottomrule
  \end{tabular}
\end{table}

\subsection{Comparison with SOTA}
 The ablation studies and comparision with state-of-the-art methods are shown in Table.~\ref{tab:sota}. The baseline here is our re-implementation of TransReID. With grayscale patch replacement (GS), the mAP and cmc1 are improved by 0.8\% and 0.5\% respectively. Combined with GS, hard batch mining (HM) provides extra 0.5\% mAP and 0.6\% cmc1, and adversarial scene removal module (SR) provides extra 0.2\% mAP and 0.6\% cmc1 gains. Compared with SOTA methods, the proposed method provides overall 0.7\% mAP and 0.8\% cmc1 improvements. Further, with re-ranking, our method achieves 84.4\% mAP and 89.9\% cmc1, surpassing all of the previous methods and setting state-of-the-art result.

\section{Conclusion}
\label{sec:formatting}
In this report, we propose hard batch mining for ReID to mine the globally similar classes to make harder mini-batches, which helps to improve the discrimination of similar classes. Besides, a scene removal module composed of a scene classifier and an adversarial loss is used to learn scene invariant feature representations. Experiments are conducted on dataset MSMT17 to prove the effectiveness, and our method surpasses all of the previous methods and sets state-of-the-art result. More datasets will be evaluated in the future.

 
{\small
\bibliographystyle{ieee_fullname}
\bibliography{egbib}
}

\end{document}